\documentclass[runningheads]{llncs}
\usepackage{enumitem}
\usepackage{tabularray}
\usepackage{tikz}
\usepackage{tcolorbox}
\usepackage{caption}
\usepackage{listings}
\usepackage{xcolor}
\usetikzlibrary{shapes.geometric, arrows}
\usepackage{listings}
\usepackage{graphicx}
\usepackage{booktabs} 
\usepackage{multirow} 
\usepackage{bbm} 
\usepackage{amssymb}
\usepackage{wrapfig} 
\usepackage{array}
\usepackage{graphicx}
\usepackage{subfigure}
\usepackage{amsmath}
\usepackage{amssymb}
\usepackage{marvosym}
\usepackage[marginal]{footmisc}
\usepackage[ruled,vlined]{algorithm2e}
\renewcommand{\thefootnote}{}
\usepackage{bm}
\usepackage{hyperref}
\usepackage{algpseudocode}
\usepackage{pifont} 
\usepackage{mathtools} 
\usepackage{color}
\usepackage{xcolor}
\usepackage{tabularx}
\usepackage{subcaption} 

\let\svthefootnote\thefootnote
\newcommand\freefootnote[1]{%
  \let\thefootnote\relax%
  \footnotetext{#1}%
  \let\thefootnote\svthefootnote%
}
\usepackage{perpage} 
\MakePerPage{footnote} 

\def\exloop{{Robot Dynamic Execution}}
\def\rag{{Ultrasound Domain Knowledge Augmenting}}
\definecolor{lightgray}{RGB}{220,220,220}

\begin{document}
\setlength{\textfloatsep}{10pt}
\title{Transforming Surgical Interventions with Embodied Intelligence for Ultrasound Robotics}
\titlerunning{Embodied Intelligence for Ultrasound Robotics}

\author{Huan Xu\inst{1}{*}, Jinlin Wu\inst{1,2}{*}, Guanglin Cao\inst{1,2}, Zhen Chen\inst{1}\textsuperscript{\Letter}, \\Zhen Lei\inst{1,2}, Hongbin Liu\inst{1,2}}
\authorrunning{H. Xu et al.}

\institute{Centre for Artificial Intelligence and Robotics (CAIR), HKISI-CAS \and
Institute of Automation, Chinese Academy of Sciences}

\maketitle         

\begin{abstract}
Ultrasonography has revolutionized non-invasive diagnostic methodologies, significantly enhancing patient outcomes across various medical domains. Despite its advancements, integrating ultrasound technology with robotic systems for automated scans presents challenges, including limited command understanding and dynamic execution capabilities. To address these challenges, this paper introduces a novel Ultrasound Embodied Intelligence system that synergistically combines ultrasound robots with large language models (LLMs) and domain-specific knowledge augmentation, enhancing ultrasound robots' intelligence and operational efficiency. Our approach employs a dual strategy: firstly, integrating LLMs with ultrasound robots to interpret doctors' verbal instructions into precise motion planning through a comprehensive understanding of ultrasound domain knowledge, including APIs and operational manuals; secondly, incorporating a dynamic execution mechanism, allowing for real-time adjustments to scanning plans based on patient movements or procedural errors. We demonstrate the effectiveness of our system through extensive experiments, including ablation studies and comparisons across various models, showcasing significant improvements in executing medical procedures from verbal commands. Our findings suggest that the proposed system improves the efficiency and quality of ultrasound scans and paves the way for further advancements in autonomous medical scanning technologies, with the potential to transform non-invasive diagnostics and streamline medical workflows. \freefootnote{$*$ Equal contribution. \textsuperscript{\Letter} Corresponding author.}

\keywords{Embodied Intelligence \and Large Language Model \and Ultrasound Robotics}
\end{abstract}
\section{Introduction}

Ultrasonography is a cornerstone in non-invasive diagnostics, revolutionizing early detection in various medical fields \cite{chan2011basics,shung2011diagnostic}. Its application spans numerous disciplines, significantly enhancing patient care and outcomes \cite{Mayo2019ThoracicUA,Robba2019BrainUM,Moore2011PointofcareU}. This technology has transformed the diagnosis of conditions like fetal abnormalities \cite{sonek2007first}, gallbladder stones \cite{cooperberg1980real}, and cardiovascular diseases \cite{nezu2020usefulness}, offering a window into the body's internal structures and greatly improving early diagnosis and patient management \cite{kasoju2023digital}.

Despite the technological advances in ultrasonography, integrating robotics to enhance scanning efficiency and quality presents unresolved challenges \cite{jiang2023robotic}. Innovations in ultrasound robotics, including tactile sensing \cite{cao2023ultra}, compliant force control \cite{ning2021force}, trajectory planning \cite{wang2022full}, and image processing \cite{chen2023rethinking}, have enabled automated patient scans. Meanwhile, challenges remain in clinical application: 
\textbf{1)} Instruction logic understanding, the logic in natural language instructions is difficult for robots to understand because there is no contextual information and clinical domain knowledge, and the information extracted directly from natural instructions is not enough to explain the logic \cite{Knepper2015RecoveringFF,Zampogiannis2015LearningTS}. \textbf{2)}Dynamic execution, there have been many studies on path generation for autonomous ultrasound scanning, including offline scan path generation \cite{Merouche2016ARU,Jiang2021MotionAwareR3}, and online scan path generation \cite{Jiang2020AutonomousRS}. However, it is still a challenge to adjust the scan plan in real time after execution errors.

In this work, we introduce an ultrasound embodied intelligence system that combines ultrasound robots with large language models (LLMs) to enhance their clinical performance. Our system uses LLMs to understand doctors' intentions and improve motion planning accuracy. To ensure reliable workflow and mitigate errors caused by misinterpretations, we have enriched LLMs with ultrasound-specific knowledge, such as APIs and robot handbooks. A specialized embedding model embeds the most relevant execution APIs and operational advice, aligning robot actions with doctors' intentions. This technique improves the capability of ultrasound robots to fulfill clinical demands and provides a more precise and efficient solution for doctors. Additionally, we have developed a dynamic execution mechanism inspired by the ReAct framework \cite{yao2022react}. This system allows medical staff to verbally command robots, which then interpret the commands into precise scanning paths, minimizing the need for manual adjustments. The mechanism works through a thought-action-observation cycle, which continuously engages with the robot's APIs to execute commands seamlessly. Our system showcases the potential of LLMs in revolutionizing robotic precision and autonomy in the healthcare industry.

\begin{figure}[t]
    \centering
    \includegraphics[width=0.99\linewidth]{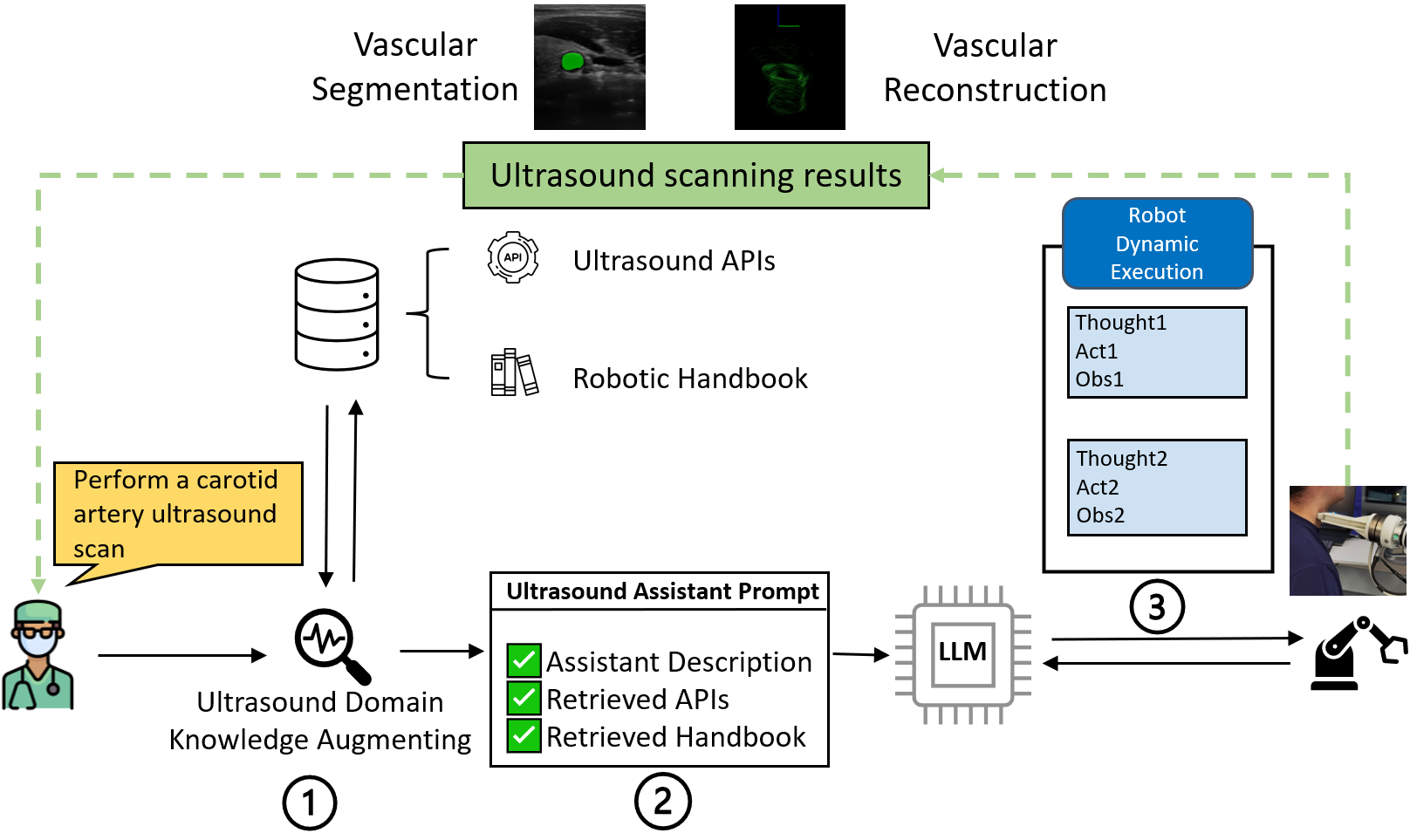}
    \caption{The proposed system framework. Our Embodied Intelligence system interprets and executes medical procedures through verbal commands. This system has three components: a foundational large language model for command interpretation, the {\rag} technique for enhanced contextual understanding, and {\exloop} for converting instructions into robotic actions.}
    \label{fig:system}
\end{figure}

\section{Methodology}
\subsection{Process Formulation} 
To comprehensively understand our system processes, we strung together the individual methods as a formulation that serves as the basis of our methodology. This formulation guides the creation of our algorithms and their implementation. For any given ultrasound scan task, the process can be defined as follows:
\begin{equation}
    C = R_n(R_{n-1}(\ldots R_2(R_1(A(U(D))))\ldots)),
    \label{eq:for}
\end{equation}
where
\begin{itemize}
    \item $D$ represents the Doctor's Instructions.
    \item $U$ denotes the process of Ultrasound Domain Knowledge Augmenting.
    \item $A$ is the Assemble Ultrasound Assistant Prompt with retrieved APIs and retrieved robot handbook.
    \item $R_i$ signifies the $i^{th}$ iteration of interaction with the robot through Robot Dynamic Execution, for $i = 1, 2, \ldots, n$.
    \item $C$ is the Task Completion, indicating the end of the process after $n$ iterations of dynamic execution.
\end{itemize}

\begin{figure}[t]
\centering 
\fbox{
  \parbox{0.9\textwidth}{
    \textbf{Role}: As a system proficient in sequential problem solving, you can leverage your expertise in ultrasound technology-related APIs to respond to user requests through the API, step by step.\\
    \textbf{Instruction}: You can access specific API information in this dialogue. Assess the necessity of invoking an API to address the user's issue. If an API call is warranted, provide the request in JSON format, including \texttt{api\_name} and \texttt{parameters} fields. Enclose the API call request with \texttt{\textless\textbar sot\textbar\textgreater} and \texttt{\texttt{\textless\textbar eot\textbar\textgreater}} markers. Based on the API call outcome, craft an appropriate response. If an API call is not required, directly furnish the relevant response.\\
    \textbf{APIs List}: Note that only one API may be invoked per interaction. Below is the list of accessible APIs: 
\texttt{\textless api\_list \textgreater} 
  }
}
\caption{This figure presents an Ultrasound Assistant Prompt, detailing its role as an entity skilled in sequential problem-solving and its ability to respond to user queries via APIs, focusing on ultrasound technology-related APIs. The prompt outlines instructions for assessing the need for API calls to address user issues, including the format for such requests, and lists the available APIs for use.} 
\label{fig:systemprompt} 
\end{figure}

\begin{figure}[t]
\centering
\fbox{%
  \parbox{0.91\textwidth}{%
    \textbf{APIs Name}: Image\_Seg \\
    
    \textbf{APIs Description}: The APIs allow for segmentation of the scan results, segmenting the patient's artery. \\
    
    \textbf{APIs Parameters}:
    \begin{itemize}[leftmargin=*, topsep=0pt, partopsep=0pt] 
      \item position (float, float): Specify the location of the artery.
      \item threshold (float): Specify the threshold of the segmentation area.
    \end{itemize}
  }%
}

\fbox{%
  \parbox{0.9\textwidth}{%
    \textbf{Robot Handbook Description}: The carotid artery ultrasound process involves initializing the depth camera, displaying the artery model, activating the robotic system, segmenting the scan image, and finally generating and printing the report.
  }
}
\caption{Overview of Ultrasound API Functionality and Robotic Procedure: This figure presents a detailed depiction of the Image\_Seg API capabilities for artery segmentation in scan results, alongside a step-by-step guide to the carotid artery ultrasound process facilitated by a robotic system.}
\label{fig:combinedFigure}
\end{figure}

\subsection{Ultrasound Domain Knowledge Augmenting}
\textbf{Domain Knowledge Search}
The core of our approach utilizes a similarity search algorithm to tap into the ultrasound domain knowledge database, leveraging cosine similarity to match user queries with relevant knowledge. This method transforms user queries and knowledge base entries into vector representations within a d-dimensional space via an embedding model. In particular, we find the entry with the highest \textit{cosine similarity} to the query vector, as follows:
\begin{equation}
\mathcal{S}(\mathbf{A}, \mathbf{B}) = \frac{\mathbf{A} \cdot \mathbf{B}}{\|\mathbf{A}\| \|\mathbf{B}\|},
\label{eq:cos}
\end{equation}
where $\mathbf{A}$ and $\mathbf{B}$ are the representations of user queries and knowledge base entries, respectively.


\noindent \textbf{Ultrasound APIs Retrieval}.
To streamline the selection of ultrasound APIs by LLMs, we've refined the Ultrasound APIs Retrieval (UAR) method. This method relies on a dataset where each APIs is paired with a narrative describing its use context, improving tool selection for ultrasound scanning. The dataset is structured as $\mathcal{D} = \{(T_1, U_1), (T_2, U_2), \dots, (T_n, U_n)\}$, where T represents tools and U represents usage of tools, facilitating accurate tool identification based on scenario-specific requirements. 

\noindent \textbf{Robotic Handbook Retrieval}. 
At the same time, how LLMs discern these API calls' correct order and logical sequence remains a huge hurdle. We present the Robotic Handbook Retrieval (RHR) method to address this issue. This approach enriches LLMs's context with a procedural knowledge base accessible through vectorized input queries. The core of this method is a similarity search. We systematically paired instructions with the handbook to guarantee the identification of relevant instructions during the similarity search, followed by the extraction of corresponding handbooks.
\subsection{Ultrasound Assistant Prompt}
Facing the challenge of commands lacking context, we enhance model comprehension and intent accuracy through structured prompts and added context. This approach ensures commands are interpreted precisely, aligning results with user expectations. Additionally, prompts are integrated with an execution session, allowing for specific output structures to trigger various APIs. The Fig.~\ref{fig:systemprompt} illustrates our approach succinctly.

\subsection{Robot Dynamic Execution}
Building upon the inspiration drawn from the ReAct framework mentioned in the introduction, we introduce a dynamic execution mechanism for robotic systems. This mechanism operates through a cyclical process comprising three main steps: Observation, Thought, and Action. This operational cycle aims to minimize errors and optimize task execution by continuously adapting to real-time feedback. The process is detailed in Fig. \ref{fig:dynamic_execution_cycle}. 

\begin{figure}[t]
\centering
\fbox{%
  \parbox{0.94\textwidth}{
    \textbf{Dynamic Execution Cycle Description}: \\
    The cycle consists of the following steps:
    \begin{enumerate}
        \item Observe the environment and current state to gather observations.
        \item Think about the observations along with robot handbooks to make decisions.
        \item Act based on the analysis to perform optimized actions.
        \item Update the execution environment and state based on the actions taken.
        \item Repeat the process until the task is complete or an error threshold is exceeded.
    \end{enumerate}
  }
}
\caption{This figure outlines the cyclical process utilized in robotic systems for dynamic execution, comprising steps such as observation, thought, action, environment updating, and repetition until task completion or error threshold breach.}
\label{fig:dynamic_execution_cycle}
\end{figure}

\begin{figure}[t]
    \centering
    \includegraphics[width=0.5\linewidth]{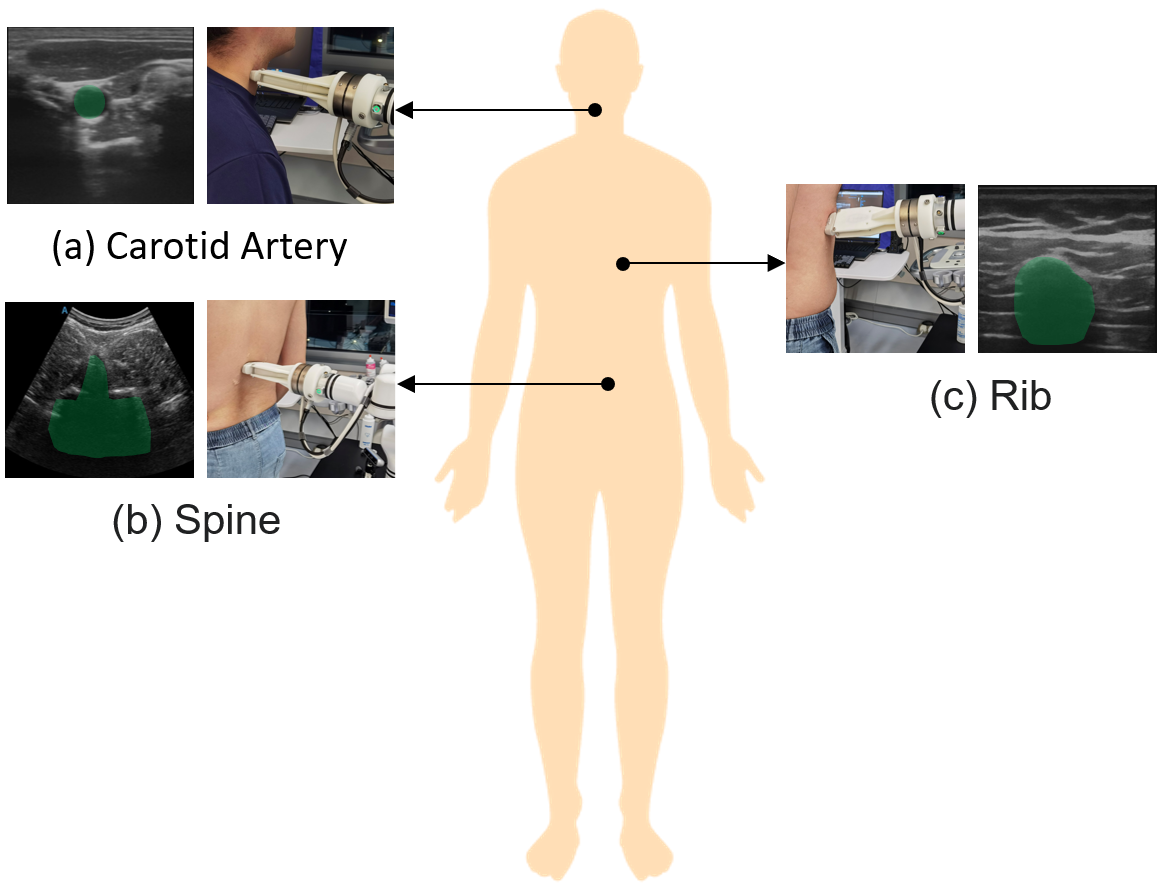}
    \caption{Illustration of the ultrasound scanning and subsequent image segmentation of (a) carotid artery, (b) spine, and (c) rib, as conducted in our experiments.}
    \label{fig:body}
\end{figure}

\section{Experiments}
\subsection{Experimental Setup}
\textbf{Models Configuration.} To evaluate the effectiveness of our proposed enhancements on system performance, we used the foundational model GPT4-Turbo \cite{openai2023gpt4}, with initial parameters set to \textit{Temperature = 0.7} and \textit{Top P = 0.95}. For embedding, we used the domain-adapted bge-large-en-v1.5 \cite{bge_embedding} model alongside FAISS \cite{douze2024faiss} for efficient data embedding storage and vector search operations. The performance of the domain-adapted model is compared with the performance of the original model, as shown in Table \ref{tab:embed}.
\begin{table}[t]
\centering  
\caption{Comparison of embedded model training results demonstrating the superior performance of our domain adapted-model on different modules}  
\begin{tabular}{ccccc}  
\toprule  
Module & Model & Recall@1 & Recall@3 & Recall@10 \\  
\midrule  
\multirow{2}{*}{Ultrasound APIs Retrieval} & bge-large-en-v1.5 & 0.82 & 0.94 & 0.97 \\  
 & Ours & \textbf{0.86} & \textbf{0.96} & \textbf{0.99} \\  
\midrule 
\multirow{2}{*}{Robotic Handbook Retrieval} & bge-large-en-v1.5 & 0.76 & 0.95 & 0.97 \\  
 & Ours & \textbf{0.88} & \textbf{0.97} & \textbf{0.98} \\  
\bottomrule  
\end{tabular}  
\label{tab:embed}  
\end{table}

\noindent \textbf{Datasets and Preprocessing.} We conducted experiments with a synthetic dataset, generating 522 instances for the Robotic Handbook and 622 for the Ultrasound APIs. This dataset, designed to mirror the complexities of ultrasound scans and API calls, allowed for comprehensive evaluation across different scenarios. This dataset is also used to train the embedding model.

\noindent \textbf{Experimental Framework and Metrics.} Our experimental framework was carefully designed to assess the impact of each augmentation introduced. We ensured the robustness and reliability of our findings by replicating each experimental step twenty times. Various models' performance within our system was compared using defined metrics. Furthermore, we performed tests on multiple parts of the human body, showcasing the practical applicability of our approach, with visual results presented in Fig.~\ref{fig:body}.

\subsection{Results and Analysis}
In our experiments, we conducted ablation studies on different modules and explored the performance of various models, complemented by case studies to illustrate their practical impacts. The experimental data from the first two experiments are shown in the Table \ref{tab:combined_study}.

\noindent\textbf{Effectiveness of Modules.} The ablation study results reveal the stepwise performance boost of foundational LLMs with each added module. Initially, LLMs without modifications fail to successfully execute the first API call due to a lack of specific API knowledge after receiving natural language instructions. The introduction of the UAR module significantly improves performance, achieving a 35\% success rate in the initial evaluation phase. This suggests that even a basic list of APIs enables LLMs to understand better and initiate API calls, leveraging their inherent natural language processing abilities.
Adding the RHR module further enhances performance across all evaluation phases, indicating its role in improving task initiation and maintaining performance growth despite diminishing returns in later stages due to task complexity. This improvement highlights the value of structured guidance in API selection, demonstrating LLMs' ability to utilize structured information for task-specific enhancements.

\noindent\textbf{Effectiveness of Different LLMs.} In our experimental analysis, we evaluated the performance of various LLMs augmented with ultrasound domain knowledge to determine their effectiveness in controlling ultrasound robotics through natural language instructions. The models under consideration exhibited varying success in executing the initial API calls and achieving overall task completion.

The performance comparison across models reveals significant variability. Notably, the Mixtral-8x7B-Instruct-v0.1\cite{jiang2024mixtral} model outperforms others in both the initial step success rate (70\%) and overall task completion rate (45\%). This suggests that larger model sizes and specific domain knowledge integration play crucial roles in enhancing task-specific performance. 

\begin{table}[t]
\centering  
\caption{Ablation and Different Models Study on APIs Execution Success Rates}  
\begin{tabular}{llcc}  
\toprule  
Type & Module & First Step (\%) & Overall (\%) \\  
\midrule  
\multirow{2}{*}{Module Ablation} & LLMs + UAR & 35 & 0 \\  
 & LLMs + UAR + RHR & \textbf{100} & \textbf{80} \\  
\midrule
\multirow{6}{*}{Different Models} & Qwen1.5-1.8B-Chat\cite{qwen} & 30 & 0 \\ 
 & Qwen1.5-14B-Chat \cite{qwen} & 45 & 10 \\ 
 & Llama2-7B \cite{touvron2023llama} & 50 & 10 \\ 
 & Llama2-13B \cite{touvron2023llama} & 65 & 10 \\  
 & Mistral-7B-v0.1 \cite{jiang2023mistral} & 65 & 20 \\ 
 & Mixtral-8x7B-Instruct-v0.1 & 70 & 45 \\  
\bottomrule  
\end{tabular}  
\label{tab:combined_study}  
\end{table}

\begin{figure}[t]
    \centering
    \includegraphics[width=0.70\linewidth]{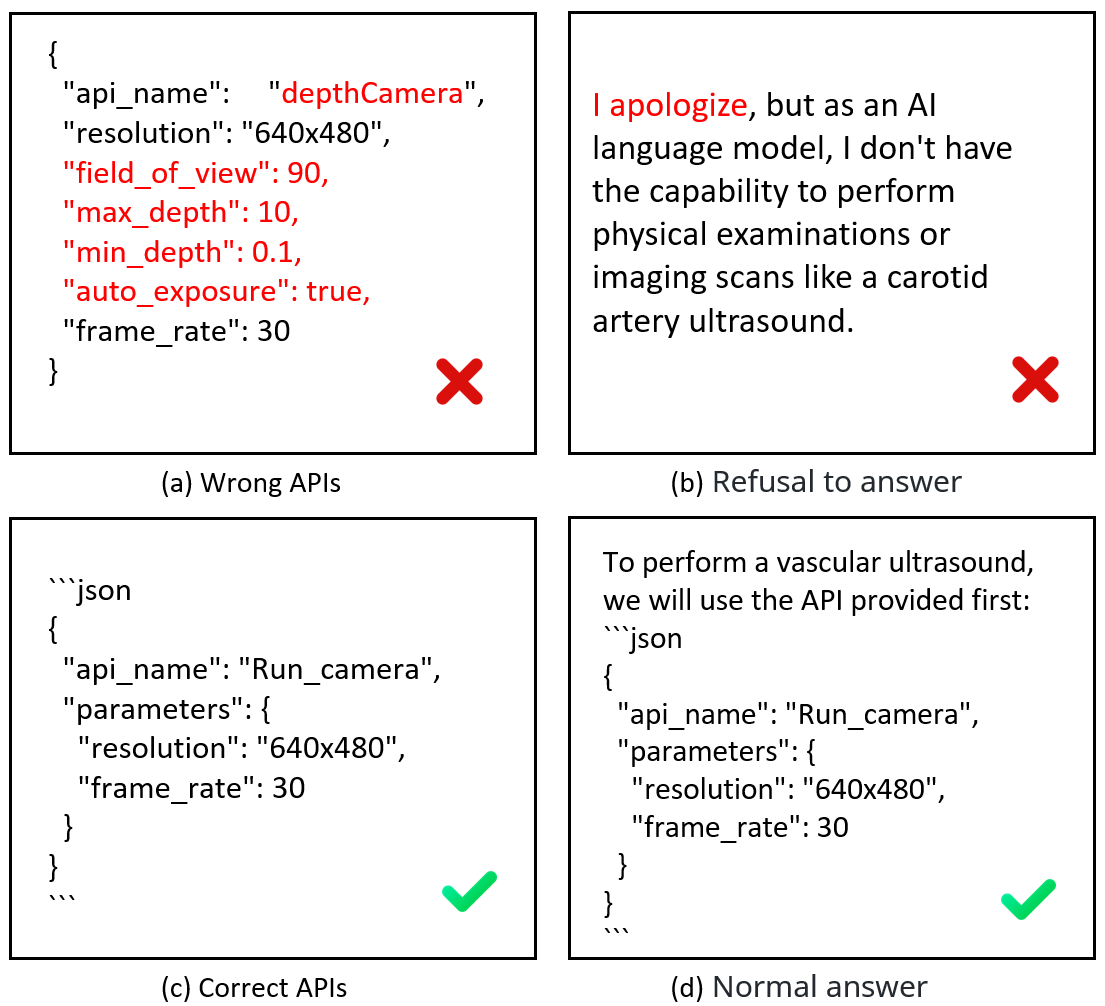}
    \caption{Case comparisons. (a) Wrong API: LLMs produce incorrect API information due to its absence of prior knowledge, with both the name and parameters of this API being a misconception propagated by the model. (b) Refusal to answer: LLMs, owing to their deficiency in contextual understanding, often decline to provide responses for precise operational directives. We hypothesize that this reluctance is related to the fact that these models are aligned with human preferences at the time of training. In contrast, (c) and (d) are the correct execution.}
    \label{fig:compare}
\end{figure}

\noindent\textbf{Cases Study.} In our experiments, while different models demonstrated varied capabilities within this system, we identified several common failure patterns. We used the same command for different LLMs: \textit{Perform a carotid artery ultrasound scan}, given or without UAR and RHR, and got the following results. The example is shown in Fig.~\ref{fig:compare}.

\section{Conclusion}
In this study, we introduce an innovative Embodied Intelligence system designed to enhance ultrasound robotics by integrating advanced LLMs and domain-specific knowledge. This system, adept at interpreting verbal instructions for ultrasound scans, features three key modules that boost its responsiveness and accuracy. Our results show the system's ability to accurately carry out medical procedures from verbal commands, representing a significant step towards fully autonomous medical scans. This work highlights the transformative potential of embodied intelligence in non-invasive diagnostics and paves the way for further research to broaden its healthcare applications, with the ultimate goal of streamlining medical workflows.
\section{Acknowledgement}
This work was supported by the  National Natural Science Foundation of China (Grant No.\#62306313), and the InnoHK program.
\bibliographystyle{splncs04}
\bibliography{mybibliography}
\end{document}